\documentclass[sigconf]{acmart}

\AtBeginDocument{%
  \providecommand\BibTeX{{%
    \normalfont B\kern-0.5em{\scshape i\kern-0.25em b}\kern-0.8em\TeX}}}

\copyrightyear{2024}
\acmYear{2024}
\setcopyright{rightsretained}
\acmConference[WWW '24]{Proceedings of the ACM Web Conference 2024}{May 13--17, 2024}{Singapore, Singapore}
\acmBooktitle{Proceedings of the ACM Web Conference 2024 (WWW '24), May 13--17, 2024, Singapore, Singapore}\acmDOI{10.1145/3589334.3648145}
\acmISBN{979-8-4007-0171-9/24/05}

\usepackage{multirow}
\usepackage{bm}

\usepackage[belowskip=2pt,aboveskip=0pt]{caption}

\begin{document}

\title{Modularized Networks for Few-shot Hateful Meme Detection}

\author{Rui Cao}
\email{ruicao.2020@phdcs.smu.edu.sg}
\affiliation{%
  \institution{Singapore Management University}
  \country{Singapore}
  \city{Singapore}
 }

\author{Roy Ka-Wei Lee}
\email{roy\_lee@sutd.edu.sg}
\affiliation{%
  \institution{Singapore University of Design and Technology}
  \country{Singapore}
  \city{Singapore}
 }

\author{Jing Jiang}
\email{jingjiang@smu.edu.sg}
\affiliation{%
  \institution{Singapore Management University}
  \country{Singapore}
  \city{Singapore}
 }
\renewcommand{\shortauthors}{Rui Cao et al.}

\begin{abstract}
In this paper, we address the challenge of detecting hateful memes in the low-resource setting where only a few labeled examples are available. Our approach leverages 
the compositionality of Low-rank adaptation (LoRA), a widely used parameter-efficient tuning technique. 
We commence by fine-tuning large language models (LLMs) with LoRA on selected tasks pertinent to hateful meme detection, thereby generating a suite of LoRA modules. 
These modules are capable of essential reasoning skills for hateful meme detection.
We then use the few available annotated samples to train a module composer, which assigns weights to the LoRA modules based on their relevance. The model's learnable parameters are directly proportional to the number of LoRA modules. This modularized network, underpinned by LLMs and augmented with LoRA modules, exhibits enhanced generalization in the context of hateful meme detection. Our evaluation spans three datasets designed for hateful meme detection in a few-shot learning context. The proposed method demonstrates superior performance to traditional in-context learning, which is also more computationally intensive during inference.

\end{abstract}





\maketitle
{\color{red} \textbf{Disclaimer}: \textit{
This paper contains violence and discriminatory content that may be disturbing to some readers.}} 

\section{Introduction}
\label{sec:introduction}
The proliferation of online social platforms has significantly enhanced the exchange and dissemination of information across communities. Memes, characterized by images paired with text overlays, have become a staple of digital communication, often designed to entertain or provoke laughter. However, the intention behind memes can take a darker turn; increasingly, they are weaponized to disseminate hate, targeting individuals or groups based on attributes like race, gender, or nationality~\cite{DBLP:conf/nips/KielaFMGSRT20,DBLP:conf/acl/PramanickDMSANC21,DBLP:conf/semeval/FersiniGRSCRLS22,DBLP:conf/emnlp/PramanickSDAN021}. Such hateful memes foster discord and may experience amplified distribution as they are repeatedly shared across various online conversations, compounding their adverse effects on social harmony. Consequently, addressing the spread of hateful memes has become a pressing issue that demands immediate attention.


\begin{figure}[t] 
	\centering
	\includegraphics[width=\linewidth]{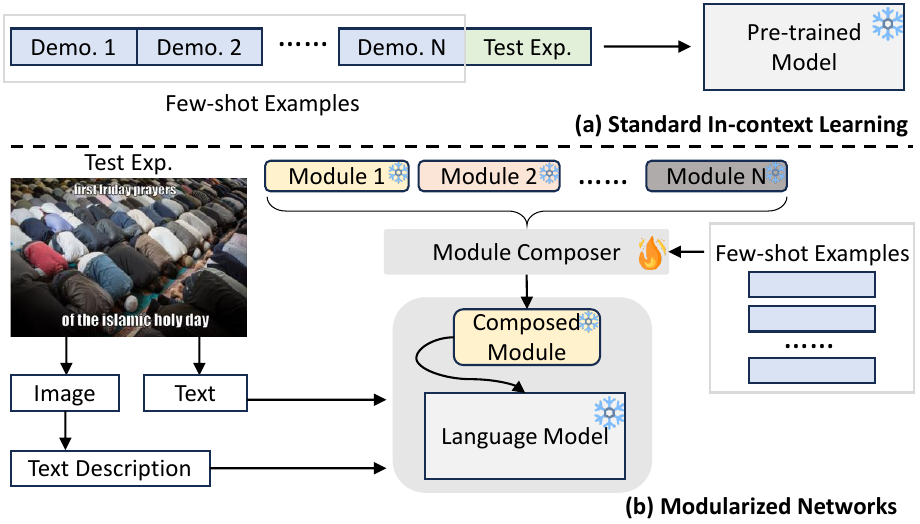} 
	\caption{Comparison of standard in-context learning and our proposed modularized networks for few-shot hateful meme detection.
        }
	\label{fig:intro-img}
\end{figure}

In response to the proliferation of hateful memes, researchers have developed various detection methods. 
One strategy regards hateful meme detection as a general multimodal classification task. It directly fine-tunes pre-trained vision-language models (PT-VLMs) to bridge the multimodal gap~\cite{zhu2020enhance,DBLP:journals/corr/abs-2012-12975,DBLP:journals/corr/abs-2012-07788,lippe2020multimodal}. Alternatively, another approach integrates these pre-trained models within specialized architectures specifically designed for detecting hateful content~\cite{DBLP:conf/mm/LeeCFJC21,DBLP:journals/corr/abs-2210-05916,DBLP:conf/emnlp/PramanickSDAN021,DBLP:conf/emnlp/CaoLC022}. 
Nonetheless, both strategies predominantly rely on extensive supervised learning, necessitating large volumes of annotated data — a process that is both costly and time-consuming. Furthermore, the emergence of hateful memes tied to evolving events poses a significant challenge; acquiring and annotating sufficient training examples for each novel occurrence is often impractical. We posit that existing studies lack adequate exploration in the low-resource setting, where the detection systems must operate effectively with minimal labeled data.


To our knowledge, the niche of few-shot detection of hateful memes has yet to be thoroughly explored in existing literature. One intuitive approach to addressing this challenge is to harness the in-context learning potential of PT-VLMs~\cite{DBLP:journals/corr/abs-2305-03726,DBLP:journals/corr/abs-2308-01390,DBLP:conf/nips/AlayracDLMBHLMM22}. This method would entail using a limited number of annotated examples as a guide for the model. For instance, when evaluating an unknown sample, a set of these annotated ``demonstration'' instances would be bundled with the test instance. This combined input is then processed by the PT-VLM to predict whether the content is hateful (as shown in Figure~\ref{fig:intro-img}(a)). While this method shows promise in low-resource scenarios for various multimodal tasks, it underperforms specifically in the domain of few-shot hateful meme detection~\cite{DBLP:journals/corr/abs-2308-01390,DBLP:conf/nips/AlayracDLMBHLMM22}. 
Additionally, the process of repeatedly combining few-shot examples with test instances incurs significant computational overhead during each inference step, which may be prohibitive in practice.


Large language models (LLMs) have recently achieved impressive results in a range of natural language processing tasks. To leverage the power of LLM in multimodal settings, a straightforward method is to convert images to captions and then feed the captions together with other textual inputs into an LLM. Previous work has adopted this method for visual question answering~\cite{DBLP:conf/emnlp/Tiong0LSH22} and robot navigation~\cite{DBLP:conf/iclr/ZengAICWWTPRSLV23}. In this paper, we also leverage LLMs for multimodal hateful meme detection following this strategy.
To tailor LLMs to new tasks, researchers have introduced various parameter-efficient tuning methods~\cite{DBLP:conf/icml/HoulsbyGJMLGAG19,DBLP:conf/iclr/HuSWALWWC22,DBLP:conf/acl/GuoRK20,DBLP:conf/acl/LiL20} as alternatives to adjusting entire models, which often consist of billions of parameters. One such method, known as Low-rank adaptation (LoRA)~\cite{DBLP:conf/iclr/HuSWALWWC22}, strategically updates weights by decomposing them into low-rank matrices, thus reducing the parameter space. 
However, LoRA is not directly applicable in our few-shot setting, due to insufficient training data.
Drawing inspiration from LoraHub's analysis of LoRA's compositional abilities~\cite{DBLP:journals/corr/abs-2307-13269}, our approach involves developing a suite of LoRA modules with essential reasoning skills for hateful meme detection. The LoRA modules are built upon LLMs, such as the LLaMA model~\cite{DBLP:journals/corr/abs-2302-13971}, and are supervised with related tasks to hateful meme detection for the acquisition of essential skills. Then, they will be combined in a modular fashion for the task of hateful meme detection.
Instead of acquiring the core competencies with a few limited examples, we transfer the learned skills from relevant tasks and only need to learn to compose these skills for hateful meme detection.

For effective detection, we have pinpointed three core competencies required: (i) Grasping the concept of hateful content; (ii) Decoding the message behind multimodal memes; (iii) Elucidating the rationale behind the hateful classification of a meme. We tailor LoRA modules to these competencies by training them on three specific tasks: (a) \textit{hate speech detection}, (b) \textit{meme comprehension}, and (c) \textit{hateful meme explanation}.
Since meme comprehension and hateful meme detection require processing multimodal data, we transform images into textual descriptions to accommodate the text-centric nature of LLMs. Subsequently, we employ supervised data from the three tasks, other than hateful meme detection, to train the corresponding LoRA modules using the LLaMA model.
With the few available examples for hateful meme detection, we train a module composer assigning importance scores over the three modules.  This composer has a minimal number of parameters, equivalent to the number of modules, making it suitable for a few-shot learning scenario. 
With the module composer, we transfer the capabilities learned from related tasks to the task of hateful meme detection through composition of LoRA modules.

By integrating the composed LoRA modules into LLaMA, we create a modular network that embodies the essential detection skills. This network contrasts with traditional in-context learning methods, notably during inference: it is more efficient as it bypasses the need to process few-shot examples with each inference step, using them solely in the training phase of the module composer.

Our validation of the proposed hateful meme detection method involved comprehensive testing across three established benchmarks. Our approach not only demonstrates greater efficiency at the inference stage
but also consistently outperforms established in-context learning baselines across all test datasets. Notably, even with a limited dataset of four examples (4-shot learning), our method outperforms the 32-shot implementation of the Flamingo model~\cite{DBLP:conf/nips/AlayracDLMBHLMM22}, which requires the entire set of examples to be processed for each evaluation instance. We summarize our contributions as follows:~\footnote{Code is available at https://github.com/Social-AI-Studio/Mod\_HATE}

\begin{itemize}
    \item We present, to the best of our knowledge, a pioneering exploration of hateful meme detection tailored to the few-shot learning setting. 
    \item We propose modularized networks with composition of LoRA modules to improve hateful meme detection. Our approach demonstrates significant efficiency improvements during the inference stage compared to traditional in-context learning models.
    \item We benchmark on three hateful meme datasets in the few-shot setting. We also perform extensive case studies to better understand the advantages and limitations of our proposed method.
\end{itemize}

\section{Related Work}
\label{sec:related}
\subsection{Hateful Meme Detection}

In the realm of online social media, memes have evolved into a prevalent mode of communication. Unfortunately, their popularity is marred by a rising trend of their use in disseminating hate. To deter this spread, numerous datasets have been developed to aid in the training of models for the detection of hateful memes~\cite{DBLP:conf/nips/KielaFMGSRT20,DBLP:conf/emnlp/PramanickSDAN021,DBLP:conf/acl/PramanickDMSANC21,DBLP:conf/semeval/FersiniGRSCRLS22}. Such detection poses a multimodal challenge, demanding an understanding of both text and image components of memes and the nuanced interactions between them. Current approaches primarily leverage PT-VLMs that are fine-tuned for this purpose~\cite{DBLP:conf/nips/KielaFMGSRT20,zhu2020enhance,lippe2020multimodal,DBLP:journals/corr/abs-2012-07788}. To better exploit PT-VLMs, some further consider ensemble of PT-VLMs~\cite{DBLP:journals/corr/abs-2012-12975,lippe2020multimodal,DBLP:journals/corr/abs-2012-07788}. However, these methods often treat hateful meme detection as a standard multimodal classification problem, overlooking the distinctive aspects of the task. Alternative strategies have sought to tailor architectures specifically for hateful meme detection~\cite{DBLP:conf/mm/LeeCFJC21,DBLP:conf/emnlp/PramanickSDAN021,DBLP:conf/emnlp/PramanickSDAN021,DBLP:journals/corr/abs-2308-08088,DBLP:journals/corr/abs-2210-05916,DBLP:conf/emnlp/CaoLC022}, yet these too rely on extensive supervised learning, which is resource-intensive and impractical for addressing the dynamic nature of memes, especially as new events unfold. Recognizing these limitations, our study introduces a novel focus on few-shot hateful meme detection, a hitherto underexplored domain. This approach is pivotal in a landscape where the rapid annotation of memes for fully supervised models is both unfeasible and cost-prohibitive~\cite{DBLP:conf/nips/KielaFMGSRT20}. By proposing an effective method tailored for the few-shot setting, we aim to bridge the research gap and provide a solution for timely and efficient detection of hateful memes.

\begin{figure*}[t] 
	\centering
	\includegraphics[width=\linewidth]{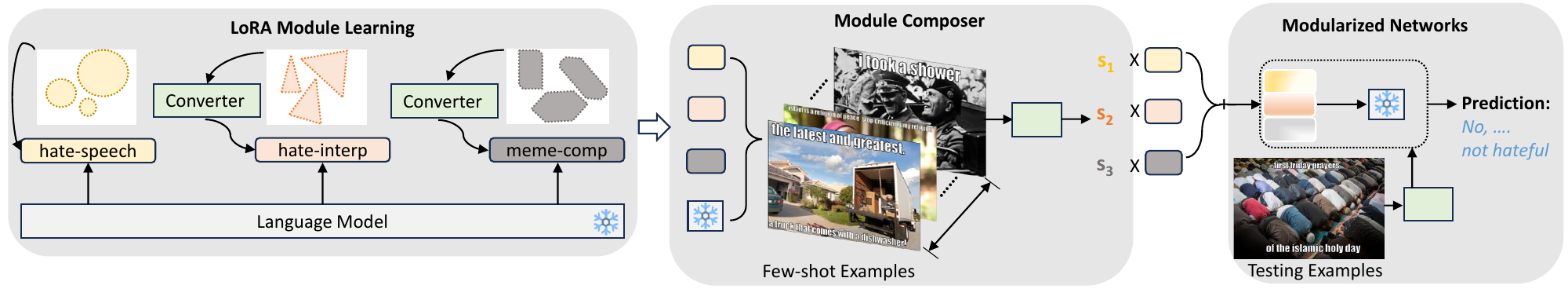} 
	\caption{Overview of the proposed Mod-HATE model. It consists of three steps: 1) LoRA module learning from relevant tasks to obtain essential skills for hateful meme detection; 2) training a module composer with few-shot training data to learn importance scores assigned over each LoRA module; 3) the construction of modularized networks by integrating the composition of learned LoRA modules with frozen LLMs.}
	\label{fig:arch-img}
\end{figure*}

\subsection{Parameter-efficient Tuning}
The adoption of large models, LLMs, has become widespread due to their impressive performance across a variety of tasks~\cite{DBLP:conf/nips/BrownMRSKDNSSAA20,DBLP:journals/corr/abs-2302-13971,DBLP:journals/corr/abs-2205-01068,DBLP:journals/corr/abs-2304-08485,DBLP:journals/corr/abs-2305-06500}. These models, however, face efficiency challenges when applied to new tasks, particularly given their multi-billion parameter scale and the typical scarcity of supervised data. To circumvent the inefficiency of full-model tuning, several parameter-efficient tuning (PEFT) techniques have been introduced. We categorize them as follows: 1) adding and updating new parameters, exemplified by prompt-tuning~\cite{DBLP:journals/ijcv/ZhouYLL22,DBLP:conf/cvpr/ZhouYL022} and adapters~\cite{DBLP:conf/icml/HoulsbyGJMLGAG19,DBLP:conf/cvpr/Sung0B22}; 2) selectively updating a sparse set of parameters~\cite{DBLP:conf/acl/GuoRK20,DBLP:conf/nips/SungNR21}; and 3) employing a decomposition strategy where updates are represented by low-rank matrices, as seen in LoRA~\cite{DBLP:conf/iclr/HuSWALWWC22}. Despite the substantial reduction in the number of parameters adjusted via PEFT methods, the challenge of tuning with very few labeled instances persists. Drawing inspiration from LoraHub~\cite{DBLP:journals/corr/abs-2307-13269}, which investigated the composability of LoRA modules and their transferability to unseen tasks, our study seeks to harness the power of LoRA for the domain of hateful meme detection. By training LoRA modules on tasks related to this domain and using a limited number of examples, we design a module composer that assigns importance scores to the learned LoRA modules. This approach enables us to tackle hateful meme detection by composing these specialized modules, aiming to effectively address the task within a few-shot learning framework.

\section{Preliminary}
\label{sec:preliminary}
 \subsection{Few-shot Hateful Meme Detection}
 \label{sec:pre-few-shot-hmd}
Given a multimodal meme (i.e., the meme image $\mathcal{I}$ and superimposed text $\mathcal{C}$), a hateful meme detection model is required to decide whether the meme is \textit{hateful} or \textit{non-hateful}. 
Models are first trained with labeled training data $\mathcal{D}_{\text{train}}= (\mathcal{I}^n,\mathcal{C}^n,\hat{a}^n)_{n=1}^N$, where $\hat{a}^n$ is the ground-truth label. Trained models are evaluated on the testing split $\mathcal{D}_{\text{test}}$.
In this work, we assume a low-resource setting where only a few labeled examples are available. We follow the definition in ~\cite{DBLP:conf/acl/GaoFC20,DBLP:conf/cvpr/ZhouYL022,DBLP:journals/ijcv/ZhouYLL22} and assume $K$ training examples per class are available in the $K$-shot setting (i.e., $N=2*K$ for our case). Our goal is to optimize the models based on $\mathcal{D}_{\text{train}}$ so that they can generalize well on the testing data $\mathcal{D}_{\text{test}}$.

\subsection{Low-Rank Adaptation}
\label{sec:lora-prelim}
LoRA~\cite{DBLP:conf/iclr/HuSWALWWC22} is a parameter-efficient tuning method, which decomposes the updates of attention weights into combination of low-rank matrices, while keeping the pre-trained weights (model parameters of LLMs) frozen. Formally, given the $i$-th attention weight matrix
 $\mathbf{W}_i \in \mathbb{R}^{p \times q}$ and its accumulated gradient update $\Delta \mathbf{W}_i$, LoRA approximates the update as follows:
 \begin{align}
     \mathbf{W}_i &:= \mathbf{W}_i + \Delta \mathbf{W}_i,\\
                & :\approx \mathbf{W}_i + \mathbf{A}_i\mathbf{B}_i,
 \end{align}
where $\mathbf{A}_i \in \mathbb{R}^{p \times r}$ and $\mathbf{B}_i \in \mathbb{R}^{r \times q}$ are the decomposed matrices with a low-rank $r$ ($r\ll p$, $r\ll q$).

LoRA largely reduces the number of trainable parameters compared with direct fine-tuning. Additionally, it allows for the composition of an LLM with various LoRA modules, each tailored to specific tasks. In this context, a LoRA module learned from one task can be considered a module with specific capabilities.

\section{Methodology}
\label{sec:method}

\subsection{Overview}
\label{sec:model-overview}
In this section, we introduce our innovative approach to few-shot hateful meme detection, known as Modularized Networks for Hateful Meme Detection (\textbf{Mod-HATE}). The core concept behind Mod-HATE is the acquisition of essential reasoning skills for detecting hateful memes through the learning of specialized modules. These modules are acquired from tasks closely aligned with hateful meme detection. 
Based on the few-shot training examples, we train a module composer to assign importance scores to these modules. Subsequently, we create a composed module by weighted averaging the learned modules. We then construct modularized networks by integrating this composed module with LLMs. The modularized networks are designed for the specific purpose of hateful meme detection. The overview of Mod-HATE is illustrated in Figure~\ref{fig:arch-img}.

Specifically, we employ LoRA based on LLMs to acquire LoRA modules with reasoning capabilities from related tasks. Since some of these tasks, including hateful meme detection, involve image information, and LLMs are inherently textual, we incorporate a converter that transforms images into textual descriptions.
The details of the converter, the introduction of relevant tasks, and the training of LoRA modules are elaborated in Section~\ref{sec:lora-modules}. In Section~\ref{sec:mod-comp}, we delve into the training of a module composer, responsible for generating importance scores for individual modules. Section~\ref{sec:modularized} is dedicated to the construction of modularized networks by the integration of composed modules with LLMs. Lastly, we demonstrate the application of modularized networks for hateful meme detection.

\subsection{LoRA Module Learning}
\label{sec:lora-modules}
In this section, we discuss the learning of LoRA modules from closely related tasks that cover essential reasoning skills for hateful meme detection.

\subsubsection{Converter}
\label{sec:converter}
As mentioned in Section~\ref{sec:model-overview}, certain tasks may involve multimodal information that is beyond the comprehension of LLMs. To address this, we employ a converter (denoted as \texttt{Converter}) to translate images into textual descriptions. It's important to note that, in line with the approach taken by authors in~\cite{DBLP:journals/corr/abs-2308-08088}, generic image captions generated by image caption generators for meme images might overlook vital cues necessary for the detection of hateful memes. Since our primary focus is not on better utilization of PT-VLMs, we have opted for an image captioning model as our converter for simplicity.
We use the PT-VLM, BLIP-2, FlanT5$_{\text{XL}}$~\cite{DBLP:conf/icml/0008LSH23} version as the captioning model. Given an image $\mathcal{I}$, \texttt{Converter} will transform it into textual descriptions, $\mathcal{T}$, of the image:
\begin{equation}
    \mathcal{T} = \texttt{Converter}(\mathcal{I}).
\end{equation}

\begin{figure}[t] 
	\centering
	\includegraphics[width=\linewidth]{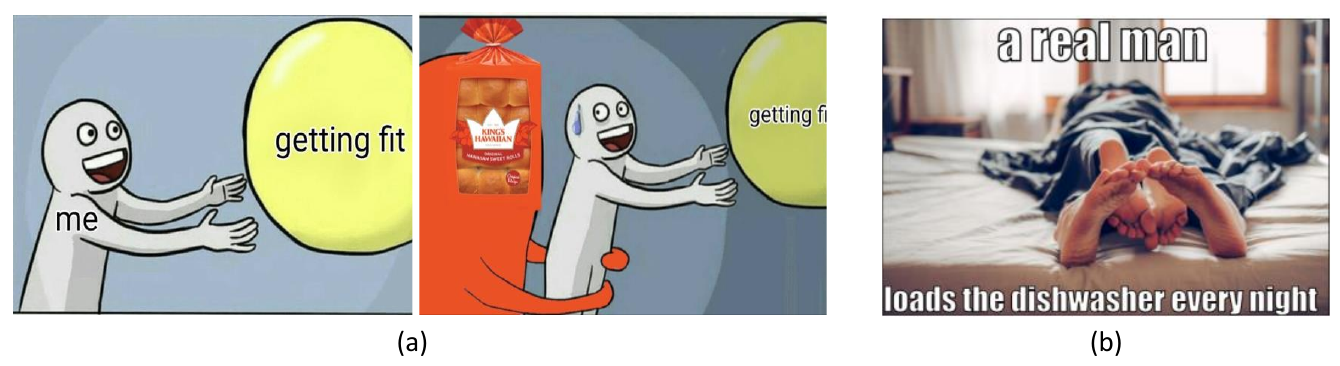} 
	\caption{Examples of (a) meme comprehension and (b) hateful meme interpretation tasks.}
	\label{fig:task-img}
\end{figure}

\subsubsection{Relevant Tasks and Supervised Data} 
\label{sec:sub-tasks}
We identify three essential skills for hateful meme detection: 1) the understanding of what constitutes \textit{hateful} content; 2) the ability to decipher the concealed meaning in multimodal memes and 3) the capability to interpret why a meme is hateful. To acquire LoRA modules proficient in these reasoning skills, we leverage three distinct tasks, each requiring one of these skills, as follows:

\noindent\textbf{Hate Speech Detection:} Given a piece of text, models are required to predict whether the text is \textit{hateful} or \textit{non-hateful}. We transform the task into a text generation task by asking the question: \textit{Please decide whether the sentence below is a hate speech. Text:} \texttt{[TEXT]}, where \texttt{[TEXT]} is a placeholder for the input text. If the text is annotated as \textit{hateful}, the expected output is \textit{Yes}, otherwise, \textit{No},. We aggregate three hate speech datasets: DT~\cite{DBLP:conf/icwsm/DavidsonWMW17}, WZ~\cite{DBLP:conf/naacl/WaseemH16} and Gab~\cite{DBLP:conf/emnlp/QianBLBW19} as the training data. To unify the datasets, we combine \textit{non-hateful} and \textit{offensive} tweets in DT dataset into the class of \textit{non-hateful}.


\noindent\textbf{Meme Comprehension:} Given a meme image $\mathcal{I}$ and the meme texts $\mathcal{C}$, meme comprehension requires to decode the meaning of multimodal memes. For instance, the meme in Figure~\ref{fig:task-img}(a) is trying to express the temptation of sweet food to the poster. To make the image comprehensible to LLMs, we generate its textual description $\mathcal{T}$ with the \texttt{Converter}. Based on the image description and the meme text, we train a module for generating the meaning of the meme with the prompt: \textit{Please interpret the meme according to its image caption and meme text. Image Caption:} \texttt{[CAP]}\textit{; Meme text:} \texttt{[MEME\_TEXT]}. The \texttt{[CAP]} and \texttt{[MEME\_TEXT]} are placeholders and will be replaced by $\mathcal{T}$ and $\mathcal{C}$ respectively. We leverage the MEMECAP dataset~\cite{DBLP:journals/corr/abs-2305-13703}, which consists of multimodal memes with their corresponding meanings, as the training dataset. 

\noindent\textbf{Hateful Meme Interpretation:} The hateful meme interpretation task requires models to give explanations as to why a meme is hateful. For instance, given the meme (i.e., the meme image $\mathcal{I}$ and the meme text $\mathcal{C}$) in Figure~\ref{fig:task-img}(b), the expected output from models should be the reasoning that the meme is annotated as hateful: \textit{the meme dehumanizes the females as sexual objects as well as less capable beings only good for dishwashing}. Similarly, as the task contains multimodal information, we leverage the \texttt{Converter} for the generation of textual image descriptions. Then we prompt LLMs with the instruction and inputs: \textit{Please explain the reason that the meme is hateful given the image caption and meme text. Image Caption:} \texttt{[CAP]}\textit{; Meme text:} \texttt{[MEME\_TEXT]}. The \texttt{[CAP]} and \texttt{[MEME\_TEXT]} will be replaced by $\mathcal{T}$ and $\mathcal{C}$ respectively. To supervise the learning of modules, we use the annotated interpretation of hateful memes in~\cite{DBLP:conf/ijcai/HeeCL23} as training data.

\subsubsection{Training of LoRA Modules}
\label{sec:training-lora-modules}
As introduced in Section~\ref{sec:sub-tasks}, we unify all relevant tasks as text generation. We adopt the widely used cross entropy loss to optimize LLMs for text generation. We use the open-source powerful language model, LLaMA (7B)~\cite{DBLP:journals/corr/abs-2302-13971} as our language model. LLaMA is a decoder-only model but we only optimize for the expected output tokens rather than both the inputs and outputs. The input tokens are masked for loss computation. Instead of tuning all parameters of LLMs, we use the LoRA parameter-efficient tuning method, introduced in Section~\ref{sec:lora-prelim} to tune the LLM. After training, we regard the set of combinations of low-rank matrices $\mathbf{A}_i\mathbf{B}_i$ as the LoRA module. Therefore, we obtain a set of LoRA modules, $\{\mathcal{L}_m\}_{m=1}^M$ regarding the considered relevant tasks, where $\mathcal{L}_m$ is the $m$-th LoRA module. In our case, $M=3$.

\subsection{Module Composer}
\label{sec:mod-comp}
Once we obtain the list of LoRA modules, the module composer will learn how to assign importance scores to these modules based on the few-shot training examples for hateful meme detection. 
The optimization objective is to generate the expected outputs for few-shot examples by composing the previously acquired LoRA modules.
Specifically, given the learned LoRA module set $\{\mathcal{L}_m\}_{m=1}^M$, our goal is to train the module composer to generate a corresponding set of importance scores $\{s_m\}_{m=1}^M$. The composed module, $\mathcal{L}_{\text{comp}}$ is computed as a weighted average over the LoRA modules:
\begin{equation}
    \mathcal{L}_{\text{comp}}=\sum_{m=1}^M s_m \mathcal{L}_m.
\end{equation}
Indeed the weighted-average of LoRA modules will be converted into the weighted-average of their low-rank matrices:
\begin{align}
    \mathbf{A}_{\text{comp},i}&=\sum_{m=1}^M s_m \mathbf{A}_i,\\
    \mathbf{B}_{\text{comp},i}&=\sum_{m=1}^M s_m \mathbf{B}_i,
\end{align}
where $\mathbf{A}_{\text{comp},i}$ and $\mathbf{B}_{\text{comp},i}$ are decomposed low-rank matrices for the $i$-th attention weight matrix's update in $\mathcal{L}_{\text{comp}}$. Next, we adapt the LLM with the composed module $\mathcal{L}_{\text{comp}}$ to construct the modularized networks. Based on the modularized networks, we optimize the module composer with both the LLM and learned LoRA modules frozen and update only the module composer with the few-shot hateful meme examples.

Similar to those multimodal tasks mentioned in Section~\ref{sec:sub-tasks} (i.e., meme comprehension and hateful meme interpretation), we use \texttt{Converter} to transform the meme image into its textual description $\mathcal{T}$. Given the meme text $\mathcal{C}$ and the meme image description $\mathcal{T}$, we ask the modularized networks the following question: \textit{Please decide whether the meme is hateful given the image caption and meme text. Image Caption:} \texttt{[CAP]}\textit{; Meme text:} \texttt{[MEME\_TEXT]}. The \texttt{[CAP]} and \texttt{[MEME\_TEXT]} will be replaced by $\mathcal{T}$ and $\mathcal{C}$ respectively. If the few-shot example is \textit{non-hateful}, the expected output will be \textit{No}; otherwise, \textit{Yes}. The modularized networks are going to optimize the importance scores (i.e., the module composer) to maximize the likelihood of generating expected outputs. We denote the loss from language modeling as $L_{\text{lm}}$. Besides, to regularize generated importance scores, L1 normalization is added to penalize extreme values. The final loss is:
\begin{equation}
    L=L_{\text{lm}} +\lambda \sum_{m=1}^M |s_m|,
\end{equation}
where $\lambda$ is a hyper-parameter adjusting the importance between the task loss and the normalization of the scores.
We adopt the same optimization strategy as introduced in~\cite{DBLP:journals/corr/abs-2307-13269} and use a gradient-free optimization method, \textit{Covariance Matrix Adaptive Evolution Strategies}~\cite{DBLP:conf/icec/HansenO96}, to minimize the loss.

\subsection{Modularized Networks}
\label{sec:modularized}
In this section, we describe the construction of modularized networks, which will be used as the architecture for training the module composer as mentioned in Section~\ref{sec:mod-comp}. The modularized networks consist of an LLM and the LoRA adapter, which is the composition of LoRA modules for relevant tasks (i.e., $\mathcal{L}_{\text{comp}}$). The attention weight matrices in the frozen LLM will be updated as:
\begin{align}
     \mathbf{W}_i &:\approx \mathbf{W}_i +  \mathbf{A}_{\text{comp},i}\mathbf{B}_{\text{comp},i},\\
                   &:= \mathbf{W}_i +( \sum_{m=1}^M s_m \mathbf{A}_i)(\sum_{m=1}^M s_m \mathbf{B}_i).
\end{align}
The networks is modularized by changing the importance scores over LoRA modules so that the importance of modules will be adjusted, to adapt to new tasks or datasets.
Based on the modularized networks, we optimize the module composer and determine the final importance scores assigned to the LoRA modules. Subsequently, we apply these modularized networks to our primary task, which is the detection of hateful memes.

\subsection{Model Prediction}
\label{sec:training}

Hateful meme detection can be conceptualized as a binary classification task. This task requires the prediction of a probability for each potential class, which is crucial for certain evaluation metrics, such as the Receiver Operating Characteristics (ROC) curve \cite{DBLP:conf/nips/KielaFMGSRT20}. From our modular networks, we extract the output probabilities corresponding to the first predicted token, denoted as $\mathbf{o}^{|\mathcal{V}|}$, wherein $\mathcal{V}$ represents the vocabulary set utilized by the LLM. For the binary outcomes of `\textit{non-hateful}' and `\textit{hateful}' memes, `\textit{No}' and `\textit{Yes}' are used as the expected outputs from the LLM, respectively. Therefore, the probability of `\textit{No}' is used to gauge the probability of the meme being classified as non-hateful. Conversely, the probability of 'Yes' is indicative of the meme being flagged as hateful. Finally, we obtain the probability of classification, $\mathbf{a} \in \mathbb{R}^2$, where $a_0=\mathbf{o}_i,  (\mathcal{V}_i=\text{No})$ and $a_1=\mathbf{o}_j, (\mathcal{V}_j=\text{Yes})$.

\section{Experiments}
\label{sec:experiment}
In this section, we first detail the evaluation framework, encompassing the datasets used, the metrics applied for assessment, and the specifics of our implementation. Subsequently, we introduce the baseline models for comparison and present our experimental results, delineating the performance contrasts between these baselines and our proposed model. Following this, we perform ablation studies to ascertain the contribution of individual components within our method. Finally, we offer case studies to elucidate the strengths of our approach, providing deeper insights into its practical application.


\begin{table}[t]
\centering
\caption{Statistical distributions of test sets.}
  \label{tab:dataset}
  \begin{tabular}{c|cc}
    \hline
    \textbf{Datasets} & \multicolumn{2}{c}{\textbf{Test}}\\
    & \#Hate. & \#Non-hate.\\
    \hline\hline
    FHM  &  247 & 253 \\
    MAMI &500 &495\\
    HarM  & 124 & 230\\
    \hline
\end{tabular}
\end{table}

\begin{table*}[t]
\centering
  \caption{Comparison with existing methods for few-shot hateful meme detection.}
\label{tab:exp-sta-comparison}
  \begin{tabular}{c|c|cc|cc|cc}
    \toprule
    \textbf{Dataset}&\textbf{\# shots} &\multicolumn{2}{c|}{\textbf{FHM}}&\multicolumn{2}{c|}{\textbf{MAMI}}&\multicolumn{2}{c}{\textbf{HarM}}\\
    \textbf{Model} && \textbf{AUC.} & \textbf{Acc.}& \textbf{AUC.} & \textbf{Acc.} & \textbf{AUC.} & \textbf{Acc.}\\
    \midrule
    OPT-13B &4  & 49.8$_{\pm3.71}$ & 50.2$_{\pm1.07}$ & 54.1$_{\pm3.31}$ & 50.0$_{\pm0.35}$ & 54.9$_{\pm7.85}$ &59.6$_{\pm3.11}$ \\
    OPT-30B &4  & 50.9$_{\pm3.00}$ & 50.0$_{\pm1.68}$ & 54.2$_{\pm4.39}$ & 50.5$_{\pm1.05}$ & 59.3$_{\pm9.19}$ & 62.3$_{\pm5.13}$\\
    OpenFlamingo-3B &4& 51.3$_{\pm1.63}$ & 49.2$_{\pm0.00}$ & 43.7$_{\pm0.51}$ & 50.3$_{\pm0.00}$ & 57.2$_{\pm1.66}$ & 35.0$_{\pm0.00}$ \\
    OpenFlamingo-9B &4& 59.4$_{\pm0.33}$ & 52.1$_{\pm0.72}$  & 59.8$_{\pm2.11}$ & 50.4$_{\pm0.90}$  &63.6$_{\pm3.15}$ & 65.2$_{\pm0.22}$ \\
    Flamingo-3B &4 & 53.6 & - & - & - & - & - \\
    Flamingo-9B &4  & 62.7 & - & - & - & - & - \\
    \cmidrule{2-8}
    OPT-13B &8  & 50.7$_{\pm3.82}$& 50.3$_{\pm1.78}$& 56.2$_{\pm4.14}$& 52.9$_{\pm3.10}$ &  61.6$_{\pm5.59}$&48.4$_{\pm8.72}$ \\
    OPT-30B &8  & 53.5$_{\pm2.61}$&51.5$_{\pm1.90}$ & 54.0$_{\pm5.56}$  & 50.7$_{\pm1.84}$& 64.2$_{\pm4.88}$ & 61.9$_{\pm6.40}$   \\
    OpenFlamingo-3B &8& 49.1$_{\pm0.44}$ & 49.2$_{\pm0.00}$  & 42.1$_{\pm1.85}$ & 50.3$_{\pm0.00}$& 59.1$_{\pm2.21}$ & 35.0$_{\pm0.00}$ \\
    OpenFlamingo-9B&8 & 58.7$_{\pm0.94}$ & 51.6$_{\pm0.52}$  & 59.1$_{\pm2.86}$ & 50.0$_{\pm0.15}$  &62.9$_{\pm2.69}$ & 65.1$_{\pm0.23}$ \\
    Flamingo-3B &8 & 54.7& - & - & - & - & - \\
    Flamingo-9B &8  & 63.9 & - & - & - & - & - \\
    \cmidrule{2-8}
    Flamingo-3B &32 & 55.3& - & - & - & - & - \\
    Flamingo-9B &32  & 63.5 & - & - & - & - & - \\
    \midrule
    \multicolumn{8}{c}{\textit{Our Proposed Method}}\\ 
    Mod-HATE&4& $\mathbf{64.5}_{\pm0.19}$&$\mathbf{58.0}_{\pm1.07}$  &$\mathbf{67.4}_{\pm0.46}$ &$61.0_{\pm2.22}$  &$\mathbf{73.4}_{\pm0.27}$ &$69.4_{\pm0.42}$  \\ 
    \cmidrule{2-8}
    Mod-HATE&8& $64.0_{\pm0.19}$&$57.4_{\pm0.82}$  &67.2$_{\pm0.15}$ &$\mathbf{61.1}_{\pm0.44}$  &$73.1_{\pm0.16}$ &$\mathbf{69.5}_{\pm0.35}$  \\ 
    \bottomrule
\end{tabular}
\end{table*}
\subsection{Evaluation Setting}
\label{sec:eval-setting}

\noindent\textbf{Datasets:} To assess the effectiveness of our proposed method, we conducted evaluations using three benchmark datasets, which are commonly used in hateful meme detection studies. The \textit{Facebook Hateful Meme} dataset (\textbf{FHM})\cite{DBLP:conf/nips/KielaFMGSRT20} encompasses a diverse range of synthetic memes, which are designed to include confounders that necessitate genuine multimodal reasoning for accurate classification, and these memes often target various vulnerable groups. The \textit{Multimedia Automatic Misogyny Identification} dataset (\textbf{MAMI})\cite{DBLP:conf/semeval/FersiniGRSCRLS22}, contains memes specifically derogatory towards women, reflecting the common targets of online vitriol. These memes are sourced from actual content on social platforms like Twitter and Reddit. Furthermore, considering hateful memes are also harmful we also utilize the \textit{Harmful Meme} dataset (\textbf{HarM})\cite{DBLP:conf/acl/PramanickDMSANC21} to test the generalizability of our method. This dataset concentrates on COVID-19 related memes and categorizes them into three levels of harm: \textit{harmless}, \textit{partially harmful}, and \textit{very harmful}. For our purposes, we have combined the latter two categories under a single label: \textit{harmful}. The statistical distributions for the original test splits of these datasets are detailed in Table~\ref{tab:dataset}.


\noindent\textbf{Evaluation Protocols:} For evaluation metrics, we employ standard accuracy (\textbf{Acc.}) and the Area Under the Receiver Operating Characteristics curve (\textbf{AUCROC}), consistent with benchmarks used in existing studies \cite{DBLP:conf/nips/KielaFMGSRT20,DBLP:conf/emnlp/CaoLC022,DBLP:conf/mm/LeeCFJC21,lippe2020multimodal,zhu2020enhance}. In the context of few-shot learning, evaluations can exhibit high variability due to the selection of sample examples. To mitigate this, we align with the approach proposed by \cite{DBLP:conf/acl/GaoFC20}, which suggests that generating multiple few-shot training sets using different random seeds can lead to a more reliable performance evaluation. We generate five sets of few-shot examples with five random seeds for each $K$-shot setting.
Consequently, we present the average accuracy and AUCROC scores computed over the test set, following training on these various few-shot samples.


\noindent\textbf{Implementation Details: }To extract the meme texts on the image, we use the open-source package EasyOCR~\footnote{https://github.com/JaidedAI/EasyOCR} for meme text detection. 
Before captioning meme images, in order to avoid the noise from the meme texts on the image, we follow~\cite{zhu2020enhance} to remove the meme texts on the image. For the optimization of the module composer, we use Covariance Matrix Adaptive Evolution Strategies~\cite{DBLP:conf/icec/HansenO96} provided by \textit{Nevergrad}~\footnote{https://github.com/facebookresearch/nevergrad}, the gradient-free optimization platform.
More details about implementations (e.g., number of model parameters, package versions and computation costs) are provided in Appendix~\ref{sec:imp-details-appendix}.

\subsection{Baselines}
\label{sec:baselines}

In this section, we introduce baselines for few-shot hateful meme detection using in-context learning. These baselines utilize a few training examples as demonstrations and prompt pre-trained models with the concatenation of demonstrations and the testing example for prediction. We examine baselines built on PT-VLMs, adept at processing combined image and text sequences for multimodal in-context learning. For instance, Flamingo, derived from the Chinchilla LLM \cite{DBLP:journals/corr/abs-2203-15556}, integrates additional parameters for multimodal pre-training, showing proficiency in various few-shot multimodal tasks. However, Flamingo is proprietary, limiting our performance evaluation to the FHM dataset using reported outcomes for its models with 3 billion (3B) and 9 billion (9B) parameters. Alternatively, OpenFlamingo, an open-source version modeled after Flamingo, uses the MPT LLM~\cite{MosaicML2023Introducing} as its foundation \cite{MosaicML2023Introducing}. We assessed OpenFlamingo with both 3B and 9B configurations. For the latter, due to the absence of multi-GPU support in OpenFlamingo, we employed Otter-9B \cite{DBLP:journals/corr/abs-2305-03726}, which is based on OpenFlamingo but further optimized for instructional tasks.

Further, large LLMs such as GPT-3 \cite{DBLP:conf/nips/BrownMRSKDNSSAA20} are also recognized for in-context learning efficacy. 
As our approach translates visual content into textual form before leveraging LLMs, for a fair comparison, we also consider in-context learning with LLMs after the image conversion.
Specifically, we employ the freely available OPT model \cite{DBLP:journals/corr/abs-2205-01068} for this purpose, a widely recognized stand-in for GPT-3. For comprehensive details on the templates used to facilitate in-context learning with these models, we direct the reader to Appendix \ref{sec:in-context-temp}.

\begin{table*}[t]
\centering
  \caption{Ablation studies of different modules in the $4$-shot setting. \textit{hate-speech} refers to the LoRA module for hate speech detection; \textit{hate-interp} is the LoRA module for hateful meme interpretation; \textit{meme-comp} is the LoRA module for meme comprehension.}
\label{tab:exp-ablations}
  \begin{tabular}{c|cc|cc|cc}
    \toprule
    \textbf{Dataset} &\multicolumn{2}{c|}{\textbf{FHM}}&\multicolumn{2}{c|}{\textbf{MAMI}}&\multicolumn{2}{c}{\textbf{HarM}}\\
    \textbf{Model} & \textbf{AUC.} & \textbf{Acc.}& \textbf{AUC.} & \textbf{Acc.} & \textbf{AUC.} & \textbf{Acc.}\\
    \midrule
    \multicolumn{7}{c}{\textit{Proposed Models  with Individual Modules, Zero-shot}}\\ 
    hate-speech& 64.3&56.0&$\mathbf{72.7}$&53.6 &$\mathbf{74.3}$&65.5\\ 
    hate-interp&56.8&49.4&56.9&50.5&60.9&35.0\\ 
    meme-captions& 47.5&51.4&46.1&48.9&40.3&64.4 \\     
    \midrule
    \multicolumn{7}{c}{\textit{Proposed Models  with Composition of Two Modules}}\\ 
    meme-comp, hate-speech& $63.3_{\pm0.17}$&$54.2_{\pm0.20}$  &69.4$_{\pm0.36}$ &$52.0_{\pm0.50}$  &$70.9_{\pm0.49}$ &$65.9_{\pm0.27}$  \\ 
    meme-comp, hate-interp& $59.5_{\pm0.07}$&$49.4_{\pm0.00}$  &55.3$_{\pm0.05}$ &$50.3_{\pm0.00}$  &$56.9_{\pm0.05}$ &$35.3_{\pm0.00}$  \\ 
    hate-speech, hate-interp& $64.1_{\pm0.40}$&$56.3_{\pm1.44}$  &67.7$_{\pm0.38}$ &$58.4_{\pm1.94}$  &$72.9_{\pm0.31}$ &$69.2_{\pm0.55}$  \\ 
    \midrule
    \multicolumn{7}{c}{\textit{Proposed Models  with all Modules}}\\ 
    Mod-HATE& $\mathbf{64.5}_{\pm0.19}$&$\mathbf{58.0}_{\pm1.07}$  &67.4$_{\pm0.46}$ &$\mathbf{61.0}_{\pm2.22}$  &$73.4_{\pm0.27}$ &$\mathbf{69.4}_{\pm0.42}$  \\ 
    \bottomrule
\end{tabular}
\end{table*}

\begin{table}[t]
\centering
  \caption{Weights of LoRA modules of our Mod-HATE model. \textbf{H-S} for the hate-speech LoRA, \textbf{H-I} for the hate-interp LoRA module and \textbf{M-C} for the meme-comp module.}
\label{tab:exp-weights}
  \begin{tabular}{c|c|ccc}
    \toprule
    \textbf{\# shots}&\textbf{Dataset} &\textbf{H-S}&\textbf{H-I}&\textbf{M-C}\\
    \midrule
   \multirow{3}{*}{4-shots}   &FHM  & $0.4865$&$0.4561$  &-0.0013 \\
   &MAMI  & $0.4210$&$0.4707$  &0.0024 \\
   &HarM & $0.4564$&0.4532&0.0025 \\
   \midrule
   \multirow{3}{*}{8-shots}   &FHM  &$0.4713$  &0.3921& $0.0013$ \\
   &MAMI  & $0.4139$&$0.4453$  &0.0014\\
   &HarM  & $0.4127$&$0.4512$  &0.0010 \\  
    \bottomrule
\end{tabular}
\end{table}

\subsection{Experiment Results}
\label{sec:exp-results-main}
Our experimental analysis was conducted under two few-shot learning scenarios: with 4-shot and 8-shot examples. The findings, as summarized in Table~\ref{tab:exp-sta-comparison}, reveal that our proposed model outperform all in-context learning baselines across all three benchmarks. These results hold true in both few-shot configurations. Furthermore, our model also demonstrates superior performance compared to the Flamingo model's 32-shot results reported in a prior study \cite{DBLP:conf/nips/AlayracDLMBHLMM22}. 



\textbf{Scaling up of models.}
Our findings suggest that increasing the size of model parameters tends to enhance the performance of in-context learning methods. This observation is in line with recent research findings~\cite{DBLP:journals/jmlr/ChowdheryNDBMRBCSGSSTMRBTSPRDHPBAI23}, indicating a positive correlation between model size and improved detection metrics. Consequently, substituting our current 7B-parameter language model, LLaMA-7B, with its larger counterparts, such as the 13B or 65B versions, could potentially lead to further advancements in performance. Such scalability suggests that our approach may be effectively adapted to larger language models, providing robust hateful meme detection capabilities with few training examples. 


\textbf{In-context learning with LLMs and PT-VLMs.}
Our study's findings suggest a marked preference for in-context learning using PT-VLMs, exemplified by OpenFlamingo and Flamingo models, over traditional LLMs such as the OPT models in the task of few-shot hateful meme detection. This difference could stem from the unique challenges that meme text and visual content interplay present, which may be particularly divergent from the data LLMs were trained on. With the limited exposure provided by few-shot examples, LLMs struggle to develop the nuanced reasoning required to decode the complex interplay of text and imagery in memes. In contrast, our method, which also utilizes LLMs, introduces a substantial enhancement by incorporating LoRA modules. These LoRA modules allow the model to master fundamental components vital for identifying hateful memes. Once these foundational skills are established, our model can adeptly adapt to the task of hateful meme detection by composing these pre-trained modules, each performs a specific reasoning task correlated to the detection task.

\textbf{Number of shots.}
Despite increasing the number of training examples, both baseline models and our proposed method exhibit a plateau in performance, with some configurations even showing a decline. For example, Flamingo-9B with 32 shots does not surpass its 8-shot counterpart, and similarly, OpenFlamingo-9B's performance does not improve when increasing from 4 to 8 shots. This suggests that the task of hateful meme detection remains challenging within a few-shot framework, and simply adding more examples does not necessarily equate to better model performance. This phenomenon indicates that the complexity of understanding and detecting nuances in hateful memes may not be adequately addressed through quantity alone. Hence, there's a clear need for more innovative approaches to effectively leverage few-shot examples. Further research might explore alternative few-shot learning techniques, more advanced model architectures that can better capture the subtleties of multimodal data, or novel data augmentation methods that enhance the model's exposure to varied examples within the constraints of few-shot learning.


\begin{table*}[!ht]
\small
\centering
  \caption{Visualization of predictions from individual modules, the compositions of two modules and from our modolarized networks. Incorrect prediction in {\color{red} red}.}
  \label{tab:case-compare}
  \begin{tabular}{|p{3.8cm}|p{4cm}|p{4cm}|p{4cm}| }
    \hline
    \textbf{Meme} & \begin{minipage}[b]{0.45\columnwidth}
		\centering
		\raisebox{-.5\height}{\includegraphics[width=\linewidth]{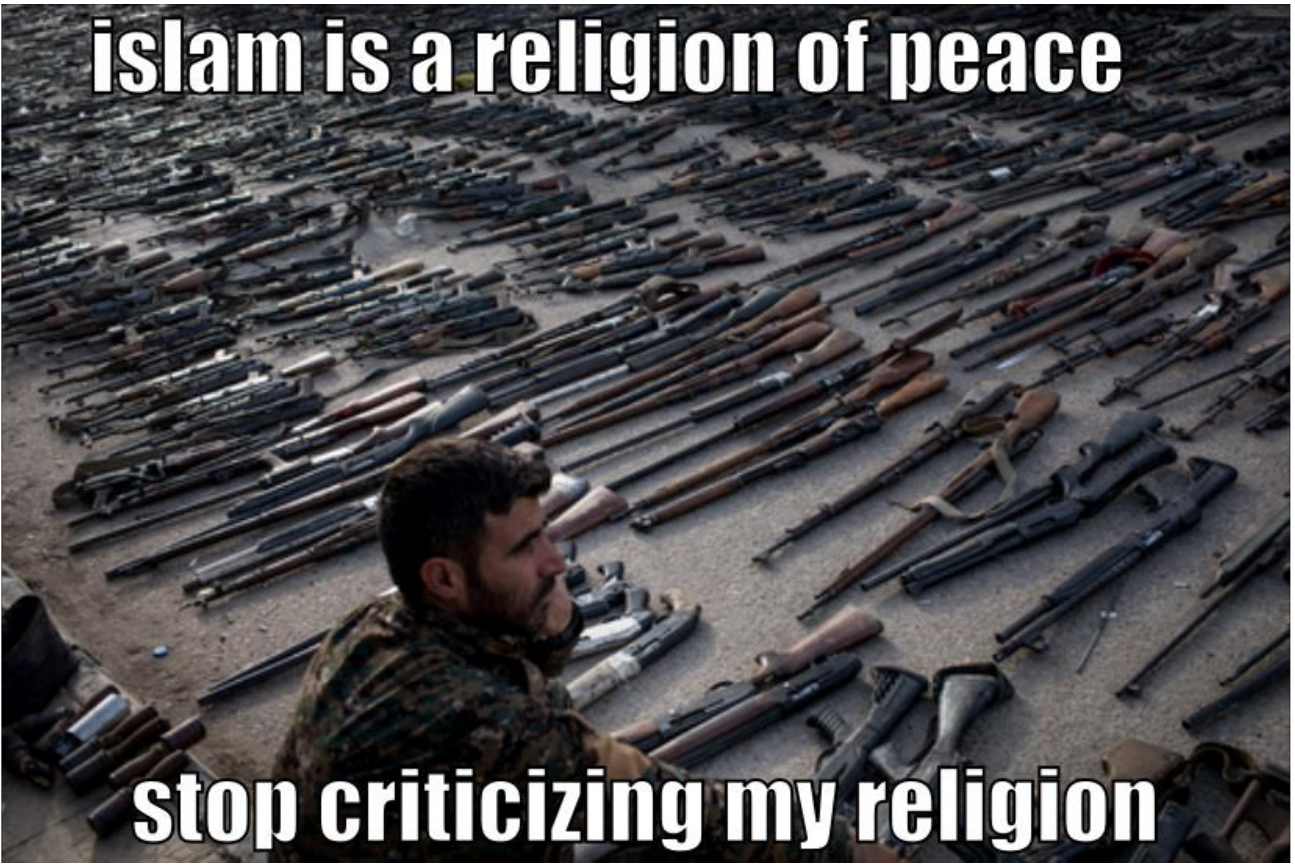}}
	\end{minipage} &
    \begin{minipage}[b]{0.45\columnwidth}
		\centering
		\raisebox{-.5\height}{\includegraphics[width=\linewidth]{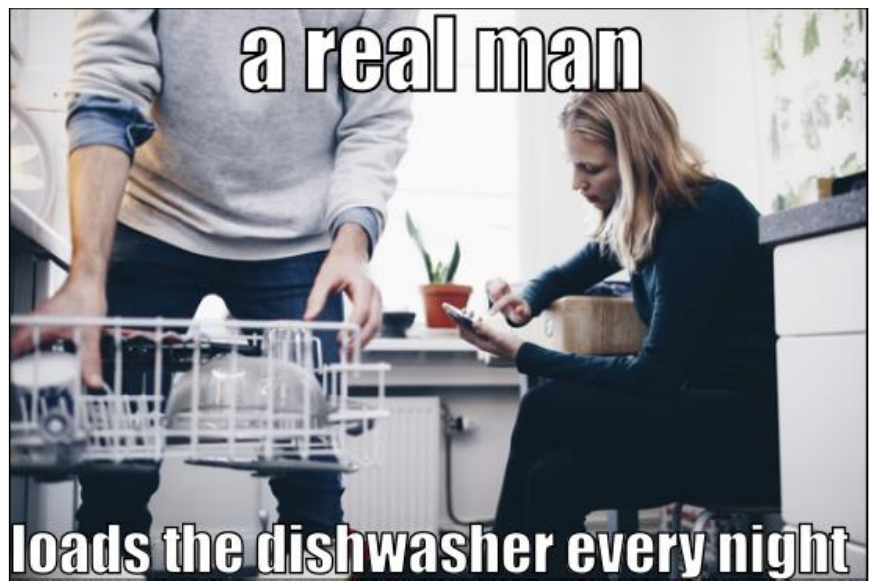}}
	\end{minipage} &
    \begin{minipage}[b]{0.45\columnwidth}
		\centering
		\raisebox{-.5\height}{\includegraphics[width=\linewidth]{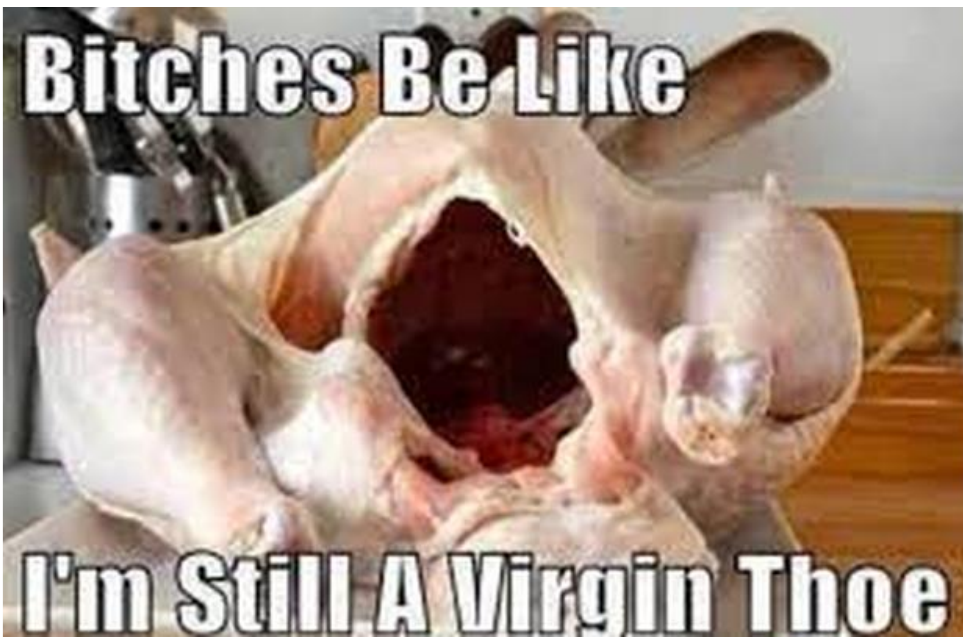}}
	\end{minipage}\\\hline
    \textbf{Ground Truth}  & Hateful (Religion) & Non-hateful& Hateful (Gender)\\\hline
    \textbf{hate-speech} & \color{red} No &No & \color{red} No \\\hline
    \textbf{hate-interp} 
    &{\color{red}{It mocks the muslims for their violent nature.}}
    &   {\color{red}{It dehumanizes the females as less capable humans that are only good for dishwashing.}}
    & {\color{red}{It insults the females by suggesting that they are only virgins because they are stupid.}}\\\hline
    \textbf{meme-comp} 
    & {\color{red}{Meme poster is trying to convey that Islam is a religion of violence.}}
    &   {\color{red}{Meme poster is trying to convey that men are supposed to do the dishes while women are supposed to do the laundry.}} 
    & {\color{red}{Meme poster is trying to convey that women who claim to be virgins are lying.}}\\\hline
    \textbf{meme-comp,hate-speech} & \color{red} No &No & \color{red} No \\\hline
    \textbf{meme-comp,hate-interp} 
    &Yes, it is hateful.
    &  \color{red} True
    &True \\\hline
    \textbf{hate-speech,hate-interp} & Yes&  No  &  \color{red} No \\\hline
    \textbf{Mod-HATE}&Yes &   No & Yes\\\hline
    \end{tabular}
\end{table*}

\subsection{Ablation Study}
\label{sec:ablation-study}

In our ablation study, focusing on the 4-shot setting due to space constraints, we examine the contributions of different modules in our modularized networks. The results, detailed in Table~\ref{tab:exp-ablations}, reveal that the integration of all three modules surpasses the performance of any two-module combination in terms of both accuracy and AUCROC, affirming the value of each module.

A noteworthy observation is the isolated hate-speech module's superior performance on three datasets regarding AUCROC. We found all its predictions to be \textit{non-hateful}.
This suggests the module learned with hate speech detection data is biased.
A plausible reason is that hate-speech module alone is incapable of understanding the interactions between meme image and meme texts (e.g., visual metaphors). 
The results underscore the necessity of the additional modules for complex multimodal understanding in broader applications. 

We further visualize importance scores of modules, as shown in Table~\ref{tab:exp-weights}. The scores do not reach the maximum due to our L1 normalization penalty that mitigates instability by avoiding extreme values in importance scores. Interestingly, even though the meme-comprehension module registers the lowest importance score, its absence negatively impacts performance. This indicates a possible underdevelopment of this module, likely due to the low resolution and noise in online-sourced memes. Enhancing the meme-comprehension module's capability through improved training on challenging data could potentially increase its contribution and overall model performance.

\subsection{Qualitative Analysis}
\label{sec:qualitative-analysis}
In this section, we conduct case studies to better understand the strengths and limitations of our proposed method.

\noindent\textbf{Case Study} Table~\ref{tab:case-compare} shows the predictions from individual modules, the composition of two modules, and the full Mod-HATE model of three example memes. The case studies shed light on the nuanced role that each module plays in both the identification and interpretability of hateful content within memes. It is noted that the modules dedicated to hateful meme interpretation and meme comprehension contribute significantly to the model's accuracy. These components not only aid in detection but also enhance the model's explanatory power (e.g., the outputs from hate-interp and meme-comp in the first example), offering a window into the model's decision-making process.

However, when these modules are used in isolation, their effectiveness diminishes, as they tend to provide descriptive explanations of the content rather than clear-cut classifications. This aligns with their underperformance when they stand alone, reinforcing the idea that the integration of modules is crucial for optimal functioning. A particular bias is detected in the hate-interp module (e.g., the second example), which has a propensity to incorrectly interpret content as hateful due to its training on exclusively hateful examples. This issue is mitigated when the module operates within the integrated framework of the Mod-HATE model, balancing out its predispositions.

The example also points out a challenge in the hate-interp module's ability to generalize to the diverse and often visually complex memes encountered in real-world scenarios, as shown in the third example. In contrast, the meme-comp module, which is trained on actual social media data, displays a more refined understanding of such content, including memes laced with visual metaphors. The hate-speech module's efficacy appears limited to situations where the cross-modal reasoning required is straightforward, struggling otherwise with more intricate multimodal interactions (e.g., the first and the third example).

In summary, the case study reveals that while individual modules possess their own strengths and limitations, their amalgamation leads to a synergistic improvement in the model’s performance. This composite approach not only bolsters the model's detection capabilities but also augments its interpretability, offering a more comprehensive solution to the challenge of hateful meme detection.

\section{Conclusion}
\label{sec:conclusion}
In this paper, we study the problem of hateful meme detection in the few-shot setting, where only a few labeled training examples are available. We propose a modularized networks which train a set of modules capable of relevant tasks to hateful meme detection and learn a composition of modules with the few-shot examples. Compared with standard in-context learning for few-shot hateful meme detection, our proposed method is more efficient as the few-shot examples will not serve as inputs during inference time which greatly reduces the computation costs. Our proposed method also outperformed all previous in-context learning on three benchmarks, demonstrating the effectiveness of the proposed method.

\section*{Acknowledgement}
This research was supported by the Ministry of Education, Singapore, under its Academic Research Fund Tier 2 (Grant No.: T2EP20222-0047, Project ID: MOE-000440-00).  Any opinions, findings and conclusions or recommendations expressed in this material are those of the authors and do not reflect the views of the Ministry of Education, Singapore.

\clearpage
\bibliographystyle{ACM-Reference-Format}
\balance
\bibliography{ref}

\clearpage
\section*{APPENDIX}
\appendix
\section{Details of Implementation}
\label{sec:imp-details-appendix}
\begin{table}[ht]
\centering
\caption{Hyper-parameters for LoRA module learning. Lr. is for learning rate, Bz. is for batch size and epoch is for the number of training epochs}
  \label{tab:hyper-params}
  \begin{tabular}{c|ccc}
    \toprule
    \textbf{Module} & \textbf{Lr.} & \textbf{Bz.} & \textbf{Epochs} \\
    \midrule
    hate-speech &0.0005 &16 &1 \\
    meme-comp &0.0005 &8 &2 \\
    hate-interp &0.0005 &8 &2 \\
    \midrule
\end{tabular}
\end{table}
We implement all models under the PyTorch Library with the CUDA-11.2 version. We use the NVIDIA A40 GPU, each with a dedicated memory of $48$GB. For the implementation of the OpenFlamingo model, we took the code released by the authors~\cite{DBLP:journals/corr/abs-2308-01390}~\footnote{https://github.com/mlfoundations/open\_flamingo}. For the implementation of LLaMA model, we leverage the HuggingFace Library~\footnote{https://huggingface.co/}, with the \textit{yahma/llama-7b-hf} checkpoint~\footnote{https://huggingface.co/yahma/llama-7b-hf}. The version of Huggingface is 4.33.0. For the parameter-efficient tuning with LoRA, we adopt implementation from the PEFT Library~\footnote{https://huggingface.co/docs/peft/index} of version 0.5.0. The training of LoRA modules is optimized with the Huggingface trainer. The hyper-parameters for LoRA module learning is provided in Table~\ref{tab:hyper-params}. The ranks of all LoRA modules are set to be $16$. We convert model parameters in LLaMA into binary float and it takes $21$GB dedicated GPU memory during the inference stage with our Mod-HATE. It takes about $21$GB dedicated GPU memory for training the module composer. 

\begin{table}[ht]
\centering
\caption{Number of parameters in models.}
  \label{tab:num-params}
  \begin{tabular}{c|c}
    \toprule
    \textbf{Model} & \textbf{\# Params (B)}  \\
    \midrule
    OPT-13B & 13\\
     OPT-30B & 30\\
    \hline 
    OpenFlamingo-3B & 3 \\
    OpenFlamingo-9B & 9 \\
    Flamingo-3B & 3.2 \\
    Flamingo-9B & 9.3 \\
    \hline
    Mod-HATE &7\\
    \midrule
\end{tabular}
\end{table}

For each meme image, we constrain the length of the meme text to be $25$. If the length exceeds, we will truncate the meme text. 
The number of model parameters are summarized in Table~\ref{tab:num-params}.

\section{Error Cases}
\label{sec:error-case-appendix}
In this section, we provide visualization of two representative error cases of our Mod-HATE in Table~\ref{tab:error}. The first kind of error comes from inaccurate predictions from some modules, as illustrated in the first example. Though hate-interp provides good interpretation, the other modules may contribute more to the final prediction so that the meme is predicted as non-hateful. For instance, the meme-comp module fails to understand the multimodal meme. It calls for better construction of individual modules as every module matters for the final prediction. 
The other common error is because different memes may rely on different modules for prediction while our module composer produces the same importance scores facing up to different memes. For example, the second hateful meme can be detected correctly with a hate-speech module as there is shallow multimodal understanding. Therefore, the skills from other modules may be redundant. 
This finding has also been proven by the results in Table~\ref{tab:exp-ablations} and better optimization methods for generating importance scores may be needed. A better solution is that the module composer can generate instance-dependent importance scores over modules.

\begin{table}[!ht]
\centering
  \caption{Error cases of Mod-Hate. Incorrect prediction in {\color{red} red}.}
  \label{tab:error}
  \begin{tabular}{|p{1.6cm}|p{2.8cm}|p{2.8cm}| }
    \hline
    \textbf{Meme} & \begin{minipage}[b]{0.38\columnwidth}
		\centering
		\raisebox{-.5\height}{\includegraphics[width=0.8\linewidth]{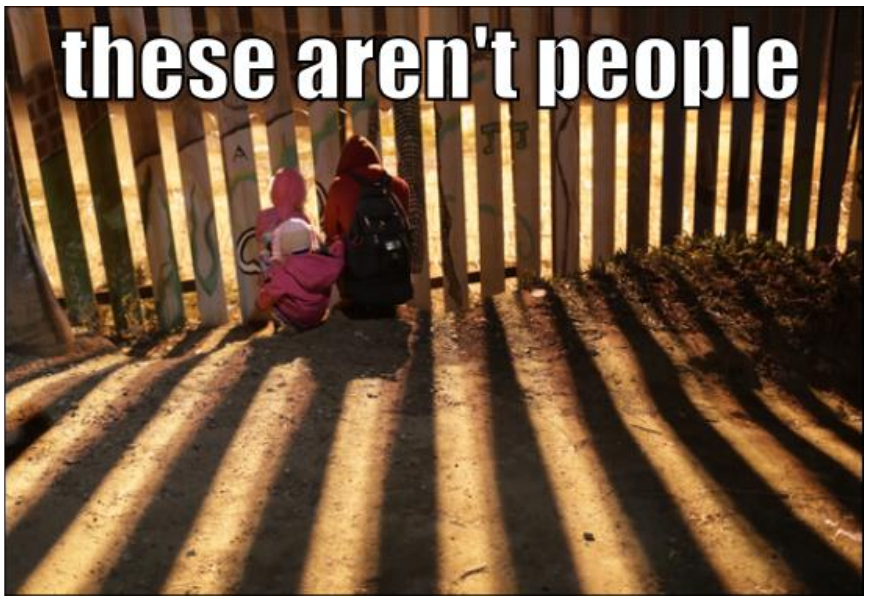}}
	\end{minipage} &
    \begin{minipage}[b]{0.38\columnwidth}
		\centering
		\raisebox{-.5\height}{\includegraphics[width=0.8\linewidth]{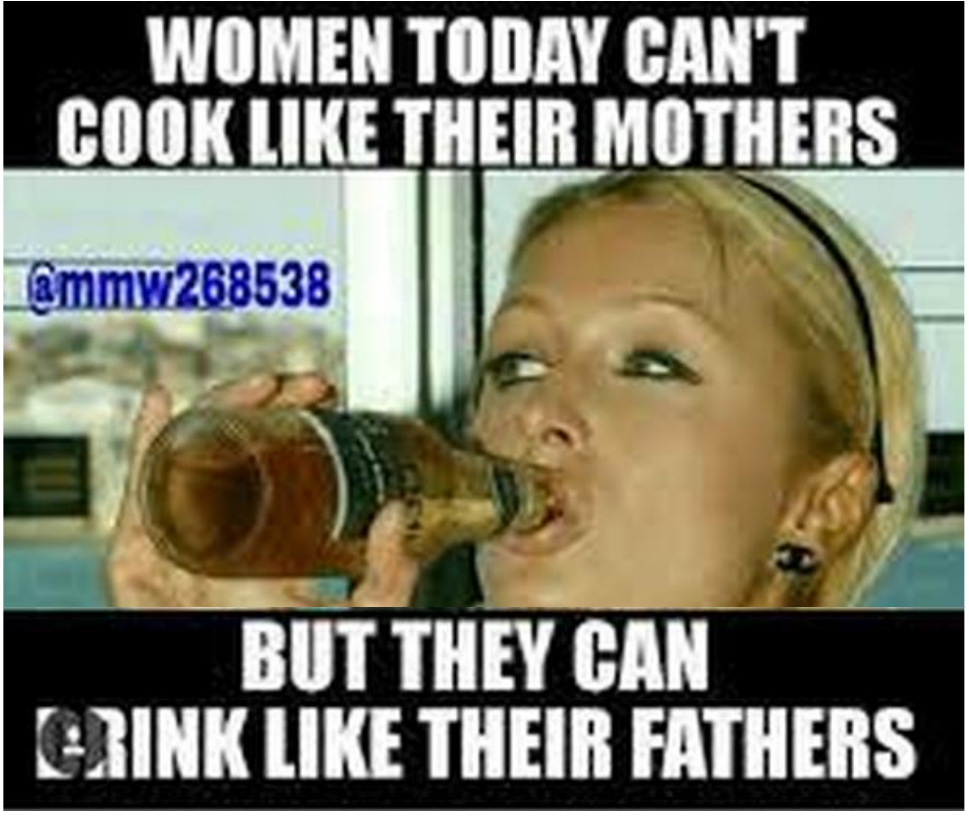}}
	\end{minipage} \\\hline
    \textbf{Ground Truth}  & Hateful (Nationality)  & Hateful (Gender)\\\hline
    \textbf{hate-speech} & \color{red} No &  Yes\\\hline
    \textbf{hate-interp} 
    &{\color{red}{It dehumanizes the immigrants as lesser humans that are not people.}}
    &{\color{red}{It dehumanizes the females as less capable humans that are only good for cooking and sex.}}  \\\hline
    \textbf{meme-comp} 
    & {\color{red}{Meme poster is trying to convey that the two children are not people but are actually a fence.}}
    & {\color{red}{Meme poster is trying to convey that women today can't cook like their mothers but they can drink like their fathers.}} \\\hline
    \textbf{meme-comp,hate-speech} & \color{red} No &  \color{red} No \\\hline
    \textbf{meme-comp,hate-interp} 
    &{\color{red}{The meme is hateful because it dehumanizes the refugees.}} 
    & {\color{red}{It is hateful because it degrades women by suggesting that they are not good cooks.}} \\\hline
    \textbf{hate-speech,hate-interp} &  \color{red} No&  \color{red} No  \\\hline
    \textbf{Mod-Hate}& \color{red} No& \color{red} No   \\\hline
    \end{tabular}
\end{table}

\begin{table*}[t]
\centering
  \caption{Weights of LoRA modules of our Mod-HATE model. \textbf{H-S} for the hate-speech LoRA, \textbf{H-E} for the hate-exp LoRA module and \textbf{M-C} for the meme-captions module.}
\label{tab:appendix--full-exp-weights}
  \begin{tabular}{c|c|ccc}
    \toprule
    \textbf{\# shots}&\textbf{Dataset} &\textbf{H-S}&\textbf{H-E}&\textbf{M-C}\\
    \midrule
   \multirow{3}{*}{4-shots}   &FHM  & $0.4865_{\pm0.0339}$&$0.4561_{\pm0.0455}$  &-0.0013$_{\pm0.0042}$ \\
   &MAMI  & $0.4210_{\pm0.0324}$&$0.4707_{\pm0.0451}$  &0.0024$_{\pm0.0043}$ \\
   &HarM & $0.4564_{\pm0.0454}$&$0.4532_{\pm0.0.0453}$  &0.0025$_{\pm0.0055}$ \\
   \midrule
   \multirow{3}{*}{8-shots}   &FHM  &$0.4713_{\pm0.0269}$  &0.3921$_{\pm0.0425}$& $0.0013_{\pm0.0013}$ \\
   &MAMI  & $0.4139_{\pm0.0102}$&$0.4453_{\pm0.0162}$  &0.0014$_{\pm0.0031}$ \\
   &HarM  & $0.4127_{\pm0.0132}$&$0.4512_{\pm0.0292}$  &0.0010$_{\pm0.0012}$ \\  
    \bottomrule
\end{tabular}
\end{table*}

\section{Templates for In-context Learning}
\label{sec:in-context-temp}
In this section, we provide the template we used for prompting pre-trained models (either PT-VLMs or LLMs) in the in-context learning manner. When prompting OpenFlamingo, we use the template: \textit{<image>User:it is an image with: }\texttt{[MEME\_TEXT]}\textit{ written on it. Is it hateful? GPT: <answer>}, where \texttt{[MEME\_TEXT]} is a placeholder and will be replaced with the real meme text. When prompting Otter, we use the template: \textit{<image>is an image with: }\texttt{[MEME\_TEXT]}\textit{ written on it. Is it hateful? Answer:}. When prompting OPT after converting meme images to textual descriptions (denoted as \texttt{CAP}), the input will be: \textit{Please decide whether the meme is hateful according to its image caption and meme text. Image Caption:} \texttt{[CAP]}\textit{; Meme Text:} \texttt{[MEME\_TEXT]} \textit{Prediction:}. The few-shot examples will be converted into similar format of prompts and are positioned at the beginning of prompts.

\section{Full Results of Importance Scores to LoRA Modules}
\label{sec:full-lora-weights}
We provide the full results of importance scores to LoRA modules in both $4$-shot and $8$-shot settings with the standard deviation in Table~\ref{tab:appendix--full-exp-weights}.

\section{Limitations}
\label{sec:limitations}
In our work, we only consider three relevant tasks to obtain skills for hateful meme detection, whereas, there may be other required skills. For instance, the comprehension of visual metaphors and figurative language is also an essential skill~\cite{DBLP:journals/corr/abs-2305-13703}. By considering a broader range of tasks, modules capable of more tasks can be obtained so that the detection of hateful meme detection can be benefited from the composition of learned LoRA modules. Secondly, we assign the same importance scores to LoRA modules given different memes, which may not be optimal. As we observe from the experiments, the detection of some hateful memes rely more on the text part. Therefore, the hate speech LoRA module plays an important role. However, others may call more for the reasoning across modalities so that depending on the hate speech module could lead to biased predictions and worse performance. In the future, we will consider an instance-dependent module composer which learns to give different importance scores considering attributes of each instance.

\end{document}